\documentclass{article}
\usepackage{spconf,amsmath,graphicx,url,txfonts}

\title{Light Pose Calibration for Camera-light Vision Systems}
\name{Yifan Song$^{\star}$ \qquad Furkan Elibol$^{\star}$ \qquad Mengkun She$^{\star}$  \qquad David Nakath$^{\star}$  \qquad Kevin K{\"o}ser$^{\star}$  } 
\address{$^{\star}$ Oceanic Machine Vision, GEOMAR Helmholtz Centre for Ocean Research Kiel, Germany}
\begin{document}
%
\maketitle
\begin{abstract}
Illuminating a scene with artificial light is a prerequisite for seeing in dark environments. However, nonuniform and dynamic illumination can deteriorate or even break computer vision approaches, for instance when operating a robot with headlights in the darkness.
This paper presents a novel light calibration approach by taking multi-view and -distance images of a reference plane in order to provide pose information of the employed light sources to the computer vision system.
By following a physical light propagation approach, under consideration of energy preservation, the estimation of light poses is solved by minimizing of the differences between real and rendered pixel intensities. 
During the evaluation we show the robustness and consistency of this method by statistically analyzing the light pose estimation results with different setups.
Although the results are demonstrated using a rotationally-symmetric non-isotropic light, the method is suited also for non-symmetric lights.
\end{abstract}
%

\section{Introduction}
\label{sec:intro}

Half of the earth's surface - specifically everything 200 meters below water level - is not illuminated by sunlight. In addition, many other environments like caves or tunnels, or cavities to be explored by endoscopy, do not provide naturally illuminated scenes. 
To successfully perform vision tasks in those conditions, artificial light sources are demanded.
To explore dark areas, an efficient way is to integrate a light source within the vision system. Thus, cars use headlights for driving in the dark and autonomous underwater vehicles (AUVs) are installed lights for exploring in the deep sea.
However, the visual appearances of objects heavily varies under changing illumination conditions and traditional computer vision solutions can struggle in such cases. 
Vision in the dark with moving light sources, especially non-isotropic ones, is not a well studied topic compared to other topics in last decades.
The knowledge about the relative pose of light sources with respect to the camera can not only improve the performance of vision based algorithms, but also facilitate many other applications like shape from shading (SFS), shape from shadow, augmented reality, photometric stereo and image-based rendering techniques in computer graphics.

\begin{figure}[t]
	\begin{center}
		\begin{minipage}[a]{0.27\linewidth}
	\centering
	\centerline{\includegraphics[width=\linewidth]{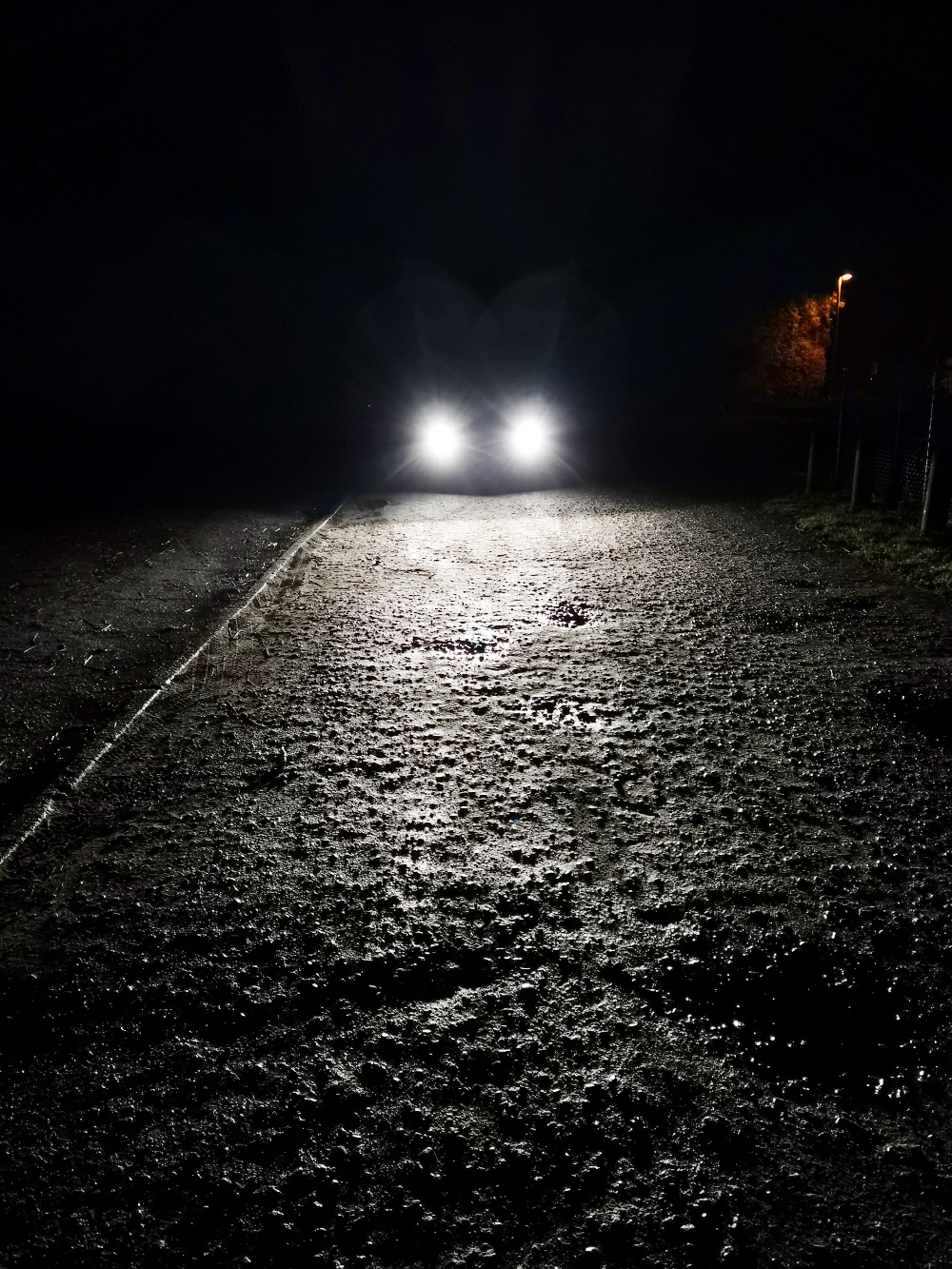}}
	\centerline{(a)}\medskip
\end{minipage}
		\begin{minipage}[a]{0.48\linewidth}
		\centering
		\centerline{\includegraphics[width=\linewidth]{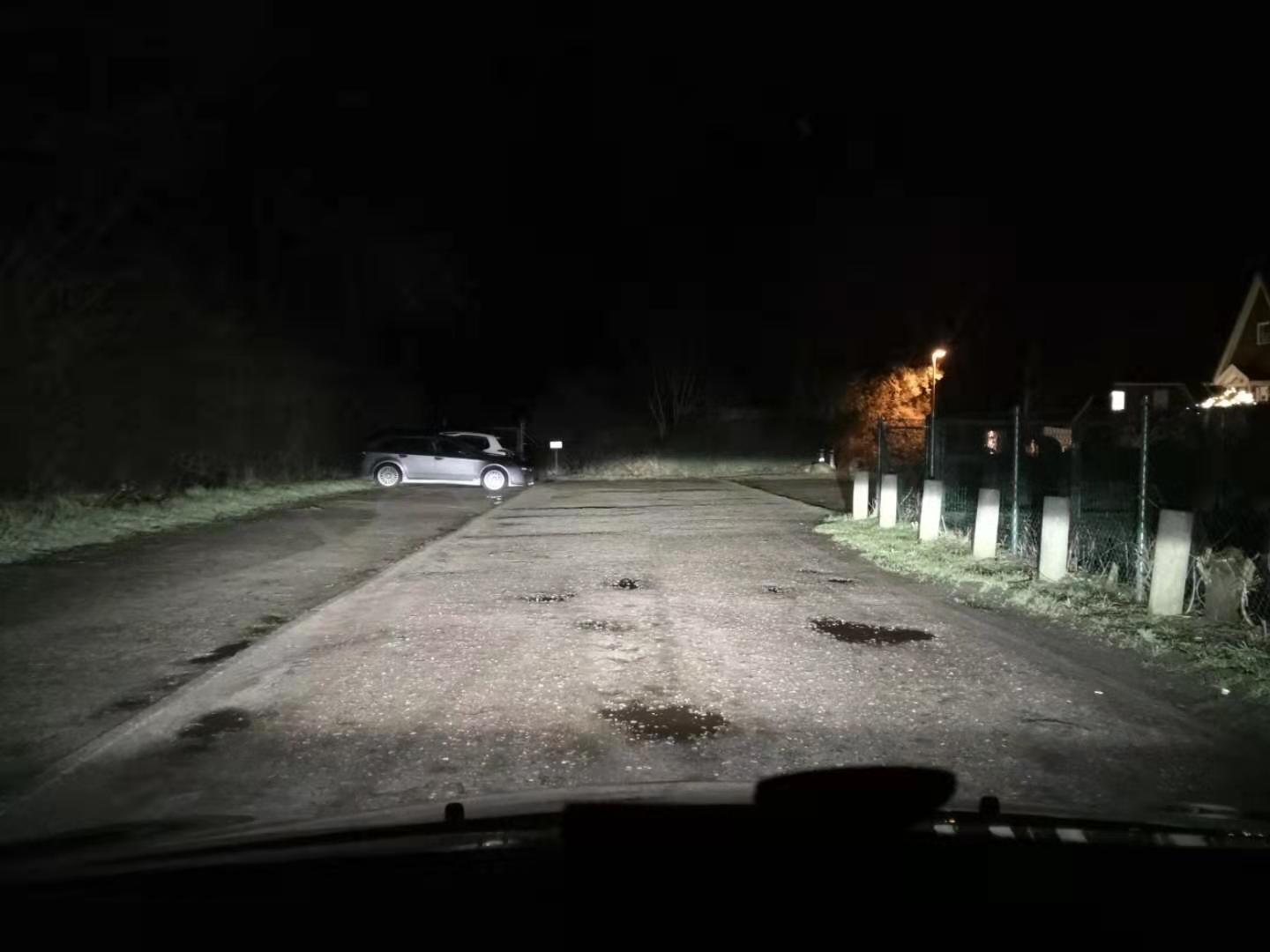}}
		\centerline{(b)}\medskip
		\end{minipage}

		\begin{minipage}[a]{0.48\linewidth}
		\centering
		\centerline{\includegraphics[width=\linewidth]{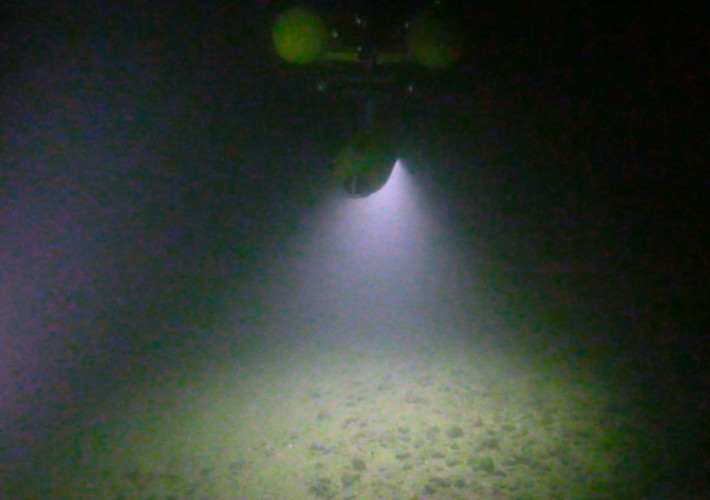}}
		\centerline{(c)}\medskip
	\end{minipage}
	\begin{minipage}[a]{0.44\linewidth}
		\centering
		\centerline{\includegraphics[width=\linewidth]{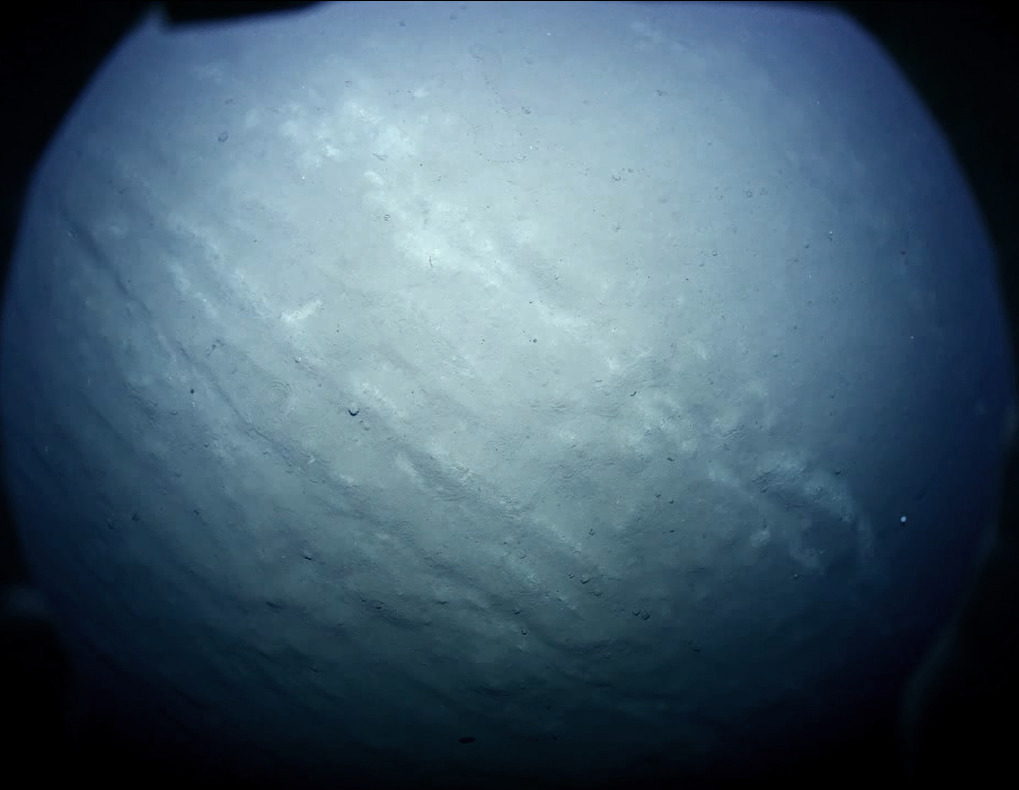}}
		\centerline{(d)}\medskip
	\end{minipage}

	\end{center}
	\caption{(a) and (b) A car is a common platform for camera-light vision system. (c) An underwater robot with co-moving light source, (d) an image taken by an underwater robot. The moving light sources challenge common computer vision algorithms.
	}
	\label{fig:intro}
\end{figure}

Hence, this paper presents a novel strategy to calibrate the fixed relative pose (position and orientation) of a directional light source (i.e. a point light with a non-isotropic angular characteristic) wrt. the camera in a camera-light vision system.
As an basis, an energy preserving rendering model is proposed and applied to estimate the relative pose parameters of lights. 
This model considers camera, object and light properties in order to render the pixel value as the irradiance which arrives on the pixel. 
The actual estimation of the relative light pose is solved by minimizing the difference between real and rendered pixel intensity values.

\section{Previous Work and Main Contributions}
\label{sec:previouswork}

The knowledge of the light pose became important in shape from shading approaches that tried to recover the 3D shape of objects according to the variations of the shading in the image.
Most of the SFS solutions assume the illuminant direction for all light rays to be parallel and that it can be estimated from  either the first derivative of the image intensity \cite{pentland1982finding}, the occluding boundary and intensity extrema \cite{yang1991sources} or the shading along image contours \cite{zheng1991estimation}. 

Another family of approaches use reflective objects to reflect the light into the image scene. The light source position can then be acquired by tracking the reflected rays from highlights in the images. The reflective objects can be specular spheres \cite{zhou2002estimation, wong2008recovering} or even a general specular surfaces \cite{jachnik2012real}. The main problem of those methods is twofold: First, the exact localization of the highlight is very difficult, since it is spatially extended. Second, because of triangulating through the reflection, highlight detection inaccuracies have a big impact on the estimated light position.

Besides detecting highlights from specular objects, different types of objects with different properties are used to infer the light source position: \cite{weber2001practical} uses a Lambertian cube to estimate the location of a light source, \cite{shen2011calibrating} designs a planar mirror, attached with a chessboard pattern and a diffuse region to recover the position of a light source and \cite{aoto2012position} uses the inside and outside highlights of a clear hollow sphere to estimate position and direction of illuminant.

Latest approaches like e.g., \cite{ackermann2013geometric} implement a light position calibration technique, which leverages Structure-from-Motion(SfM) algorithms to optimize the triangulation of reflected highlights from at least two reflective spheres in a single image. \cite{santo2018light} use a more general calibration object with a Lambertian plane and small shadow casters, to estimate the shadow caster positions and the illuminant position and direction by solving a SfM problem.
However, all methods introduced above treat light source as isotropic and only estimate the location of it. 
To the knowledge of the authors, only \cite{park2014calibrating} gives a solution to calibrate the light position and orientation for a non-isotropic point light from the multi-view images of a weakly textured planar scene. 
However, this approach is based on reflections and works only for a single rotationally symmetric light source.

The main contributions of this paper are:
(1) Propose a practical solution to calibrate the relative poses of the light sources in wrt. the camera in the camera-light vision systems.
(2) Using a physical light propagation model to simulate pixel intensities under consideration of energy preservation.
(3) Use an analysis-by-synthesis approach to solve light pose calibration by optimization.

\section{Energy preserving rendering models}
\label{sec:models}

\begin{figure}[t]
	\begin{center}
		\includegraphics[width=\linewidth]{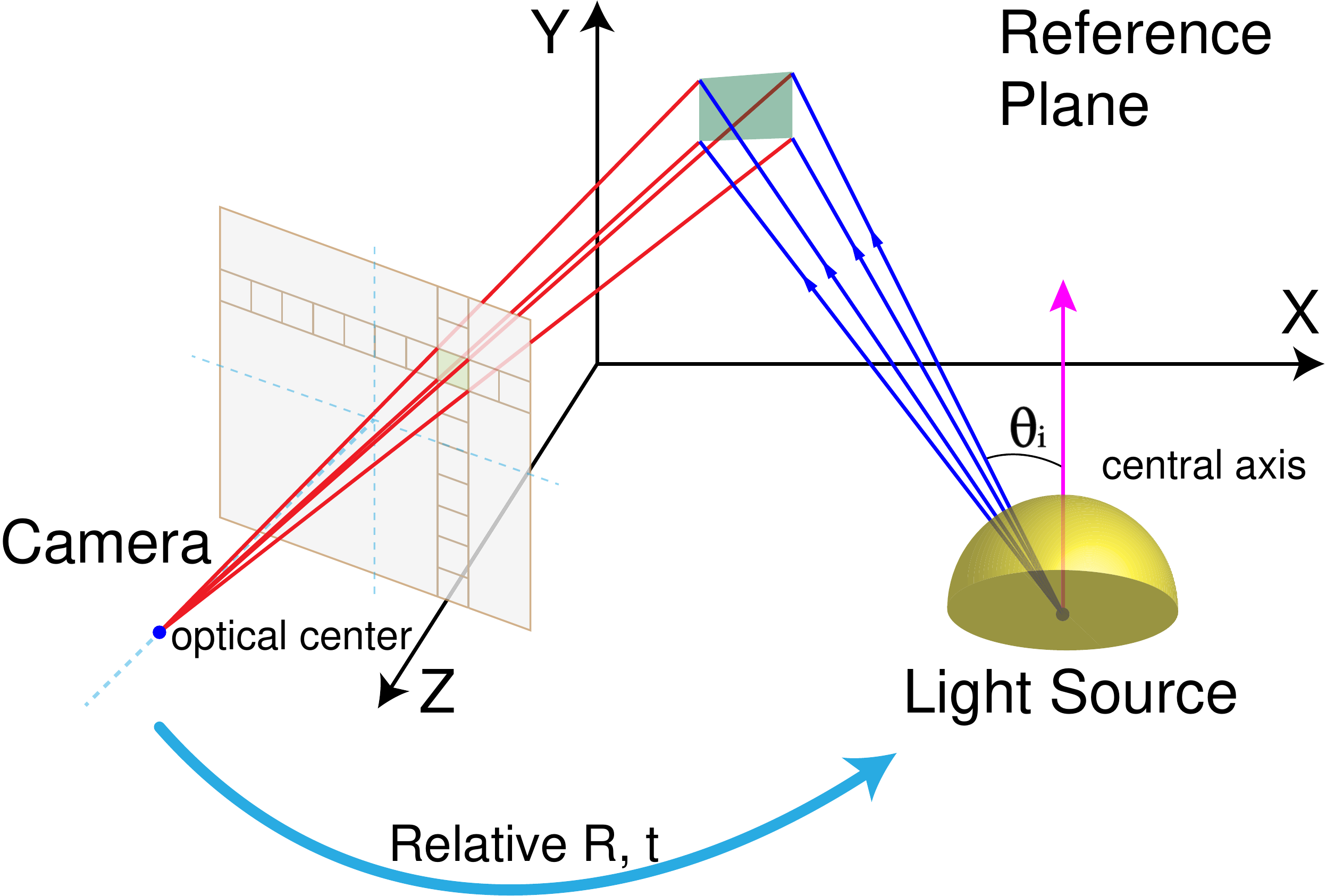}
	\end{center}
	\caption{Geometry relationships between models.}
	\label{fig:principle}
\end{figure}

This section describes the physical models involved in the rendering methods under consideration of energy preservation. 
Their corresponding geometric relationships are depicted in Figure \ref{fig:principle} and described in the following.

As we will use a reference plane (e.g. a flat white wall, which is used in this paper, or any mobile planar single color target) for calibration, we set this to be the $z=0$ plane of the world coordinate system, with the $Z$-axis parallel to the reference plane surface normal $\boldsymbol{\rm \hat{n}}$. Each pixel in the image is modeled as a square with four vertices and is back-projected to the reference plane as a quadrilateral.
We further assume that the reference plane has a Lambertian surface which reflects the incident light to all directions equally, then the intensity (illuminance) of a pixel is proportional to the energy arriving on the back-projected quadrilateral on the reference plane. 
From an energy preserving perspective, the light energy is completely distributed on a hemisphere with the light source located at its center. 
When the vertices of the quadrilateral are now projected onto the surface of such a light hemisphere, the energy passing through the thus-defined area is the same as the energy arriving in the area defined by the quadrilateral on the reference plane.

\subsection{Camera and projection model}
\label{ssec:camera}

The camera model in this paper is the perspective model. From an energy preserving point of view, pixels in CCD arrays are treated as squares constituting an area rather than infinitely small points. 
The pixel intensity is interpreted as light energy arriving on each cell of the CCD chip. The back-projected region on a reference plane for each pixel can be acquired by shooting four rays from pixel corner vertices and intersecting with the plane. Hence

\begin{equation}
\label{eqProjection}
\boldsymbol{\rm P} = \lambda\frac{\boldsymbol{\rm R}\boldsymbol{\rm K}^{-1}\boldsymbol{\rm p}} {\| \boldsymbol{\rm R}\boldsymbol{\rm K}^{-1}\boldsymbol{\rm p} \|_{2}} +\boldsymbol{\rm C}
\end{equation}

Where $\boldsymbol{\rm P}\in \varmathbb{R}^{3}$ and $\boldsymbol{\rm p}\in \varmathbb{P}^{2}$ represent the 3D and 2D coordinates of pixel corner vertices respectively. $\boldsymbol{\rm K}$ stands for the camera matrix which holds intrinsic parameters. 
In addition, distortion parameters have also been considered during the projection. 
All those parameters can be achieved from a standard camera calibration procedure \cite{Zhang1999_FlexibleCamera}. 
$\boldsymbol{\rm R}$ and $\boldsymbol{\rm C}$  are extrinsic parameters which denote the rotation and center of the camera. 
Those extrinsics can be computed by standard SfM approaches. Finally, $\lambda$ is the scale factor, when the coordinate system is based on the reference plane (world) coordinate system , it is equal to the $Z$-component of the viewing ray divided by last element of $\boldsymbol{\rm C}$.

Besides the geometric calibration of the camera, a radiometric calibration is also demanded in order to recover the linear relationship between the pixel intensity value and the light energy which has arrived on it. We obtain the response curve according to \cite{Debevec_97-hdr}.

\subsection{Reflectance model}
\label{ssec:reflection}
Reflectance rendering is a well-studied topic in computer graphics. A well known model is the bidirectional reflectance distribution function (BRDF). 
This paper adapts Lambert's Cosine Law since the filmed object, i.e. a white wall, can be considered as a Lambertian surface. Hence, we apply the Inverse Square Law to fulfill the energy preserving property.

\begin{equation}
\label{eqReflection}
I_{\boldsymbol{\rm p}} = I_{r} \frac{\boldsymbol{-\rm \hat{n}}^{\rm T} \boldsymbol{\rm l}}{d^{2}}
\end{equation}
 
In this model, light $I_{r}$ that is cast onto a surface will be reflected equally to all directions and the reflected irradiance only depends on the incident angle, which can be derived by the dot product of the incident light ray $\boldsymbol{\rm l}$ with the reference plane surface normal $\boldsymbol{\rm \hat{n}}$. The $I_{\boldsymbol{\rm p}}$ received by the camera decreases quadratically with the distance $d$ from the 3D reference plane point $\boldsymbol{\rm P}$ to the camera.

\subsection{Light source model}
\label{ssec:light}
Light sources irradiance models can generally be grouped into 2 categories: isotropic and non-isotropic. 
The latter category can continuously be classified as symmetric and arbitrary pattern of light. 
Different types of lights require different parameterizations to properly describe the relative pose with respect to the camera. 
The isotropic light model only considers the relative position of the light source as its orientation is irrelevant (same radiance to all directions). Rotationally symmetric non-isotropic light needs another two rotation angles to describe the relative rotation from the camera's optical axis to the light's central axis (rotation around central axis is irrelevant due to its symmetric property). 
The angular characteristic can be formulated as radiation intensity distribution (RID) curve,
which is assumed to be known in this contribution, but which could potentially also be included in the optimization scheme presented later.
For lights with non-symmetric angular characteristic, the radiance pattern can be stored into a grid, then this grid can be used to lookup the corresponding radiance energy; and these lights are characterized by an additional rotational degree of freedom for their pose.

The approach taken in this contribution can handle both symmetric and non-symmetric lights.
For the experiments and clarity of presentation we will however restrict ourselves to symmetric lights:
\begin{equation}
\label{eqLight}
I_{r} = s \cdot \Omega \bar{E}(\theta)
\end{equation}

Where $\Omega$ is the solid angle formed by the projected pixel vertices on the light hemisphere. 
$\bar{E}(\theta)$ denotes the average irradiance from the RID curve, which only depends on the angle $\theta$ between light ray and the light's central axis. 
Since the RID gives a relative measurement of light energy distribution, a scale factor $s$ is included to cover all scale effects (e.g. analogue-digital conversion, reference plane surface albedo) which linearly links the relative radiance measurement to pixel intensity value.

\begin{figure}[t]
	\begin{center}
		\centering
		\includegraphics[width=0.4\linewidth]{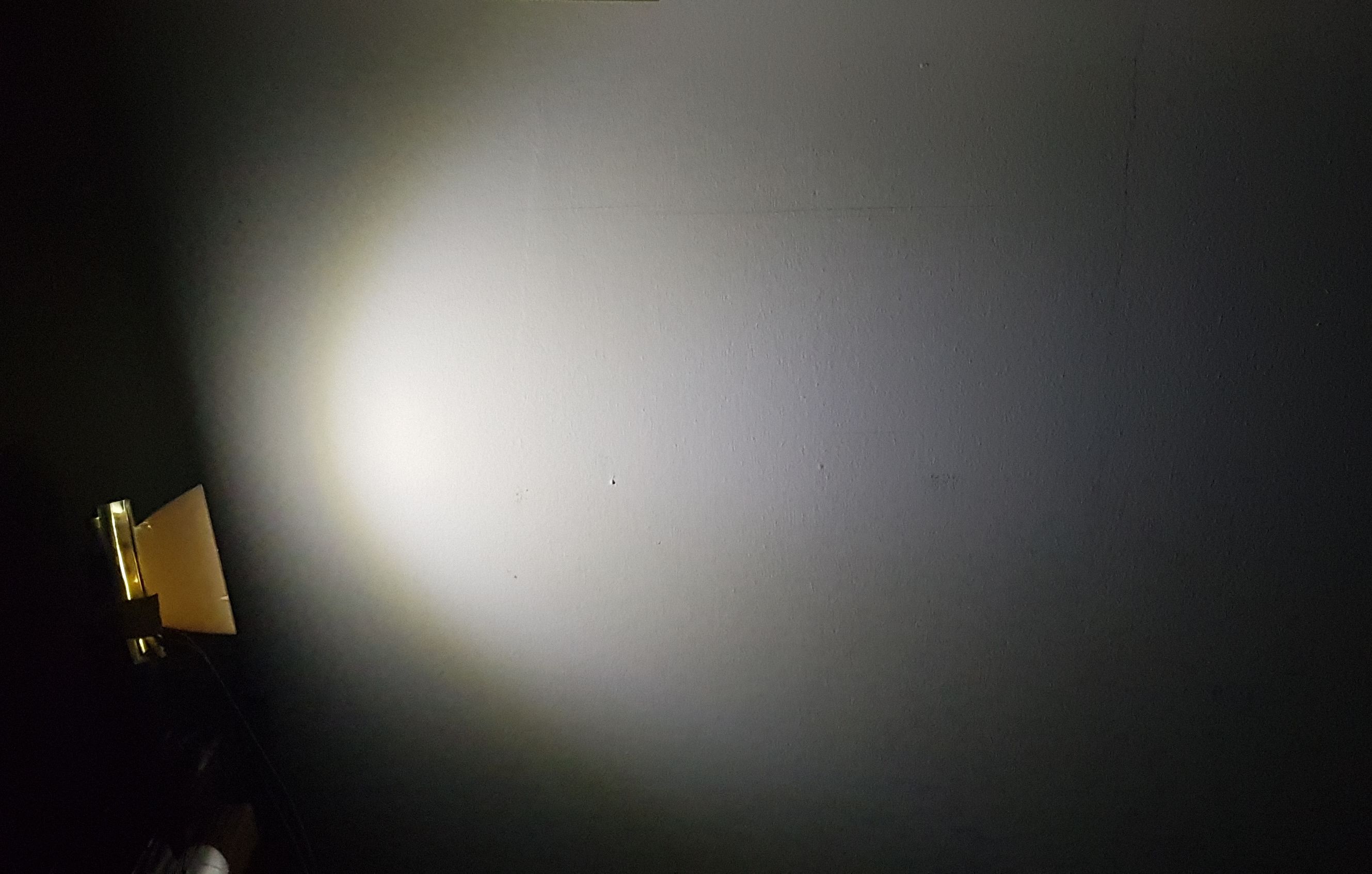}
		\hspace{0.1mm}
		\centering
		\includegraphics[width=0.55\linewidth]{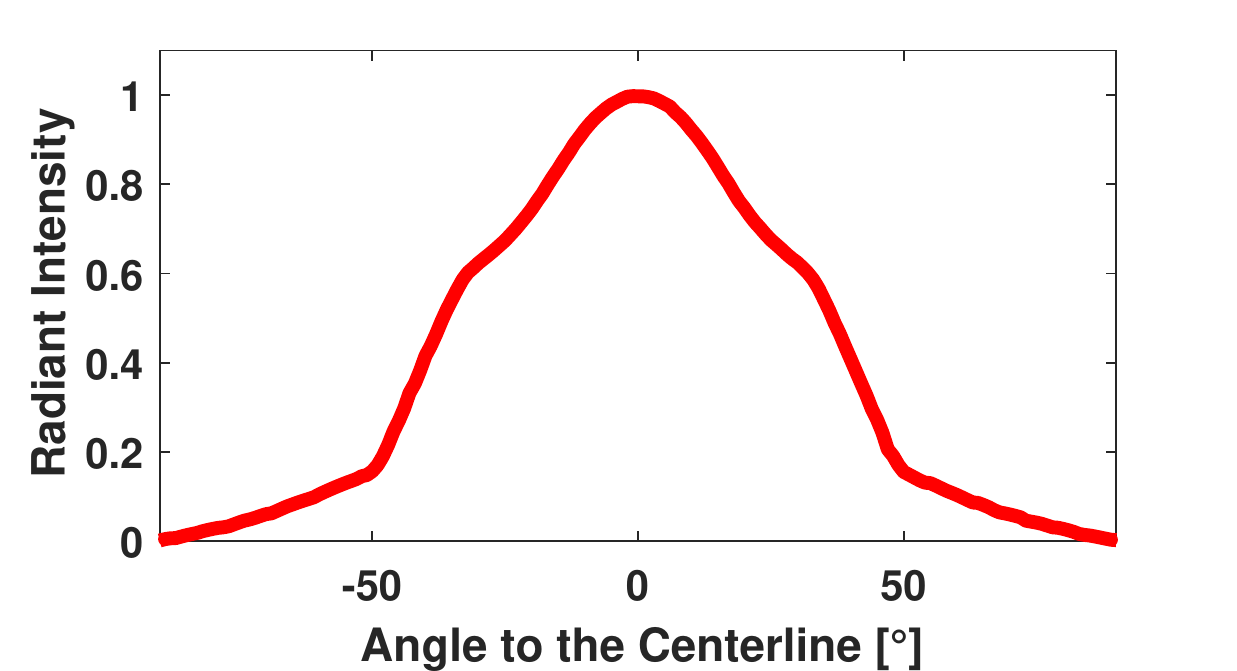}
	\end{center}
	\caption{ An example of symmetric directional light source with its RID curve.}
	\label{fig:light}
\end{figure}

\section{Methodology}
\label{sec:methodology}

This section outlines the approach to estimate the relative pose (position and orientation) of a non-isotropic point light, which is rigidly attached to a camera, by using an energy preserving rendering approach. 
All the images used in this paper are single channel raw images, which give more dynamic ranges in order to achieve higher accuracy.
Assume that several images of a flat reference plane have been taken by a camera-light vision system. 
Now, an estimation of the relative pose between camera and light can be obtained by the following steps:

\paragraph*{(1) Camera Calibration}
Geometric calibration is implemented by a standard camera calibration procedure to compute the camera matrix and distortion parameters. 
Radiometric calibration obtains the response curve of the camera, The intensity values of the acquired images can then been corrected into a linear space.
\paragraph*{(2) Multi-view SfM}
SfM is performed to obtain the extrinsics (rotation matrix and camera center) for each image in the reference plane coordinate system, alternatively markers can be used.
\paragraph*{(3) Rendering}
Select several pixels in each image and render their intensity values under the initial light poses and scale factor setups employing the energy preserving rendering model from Section \ref{sec:models}.
\paragraph*{(4) Optimization}
Minimize the difference between measured pixel intensity values $I$ and rendered ones $I_{render}$ to optimize the initial light poses (for rotationally symmetric light: light position $\boldsymbol{\rm C}_{l}$ and two Euler angles $roll_{l}, pitch_{l}$) and scale factor $s$. The optimization target function is formulated as:
\begin{equation}
\label{eqOptimization}
\delta I = \operatorname*{argmin}_{\delta I} \sum_{\boldsymbol{\rm p}} | I - I_{render}(\boldsymbol{\rm C}_{l}, roll_{l}, pitch_{l},s) | ^{2}
\end{equation}

\section{Experiments and evaluations}
\label{sec:experiments}

\begin{figure}[htb]
	\begin{center}
    \begin{minipage}[a]{0.48\linewidth}
	\centering
	\centerline{\includegraphics[width=\linewidth]{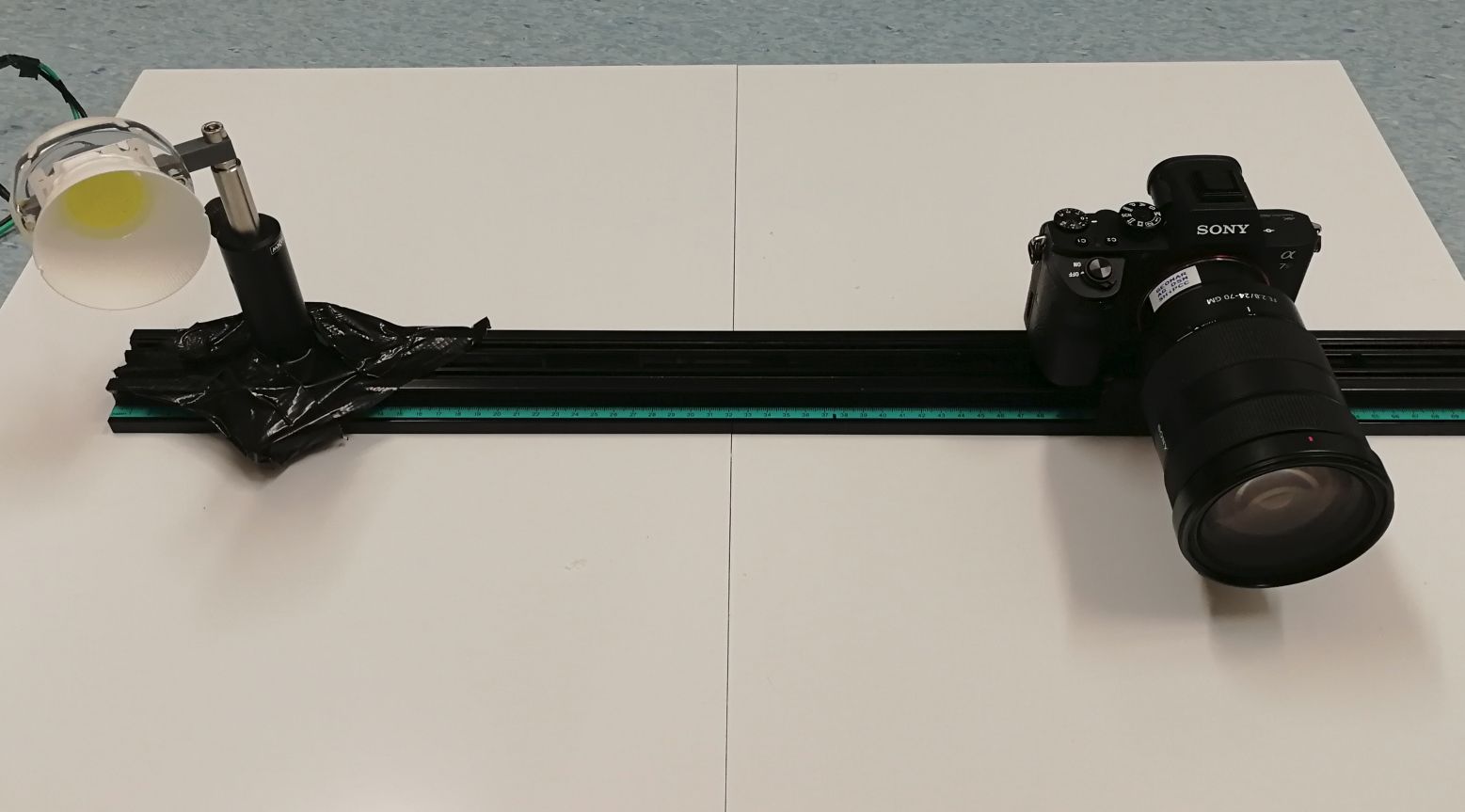}}
	\centerline{(a)}\medskip
	\end{minipage}
	\begin{minipage}[a]{0.48\linewidth}
    \centering
    \centerline{\includegraphics[width=\linewidth]{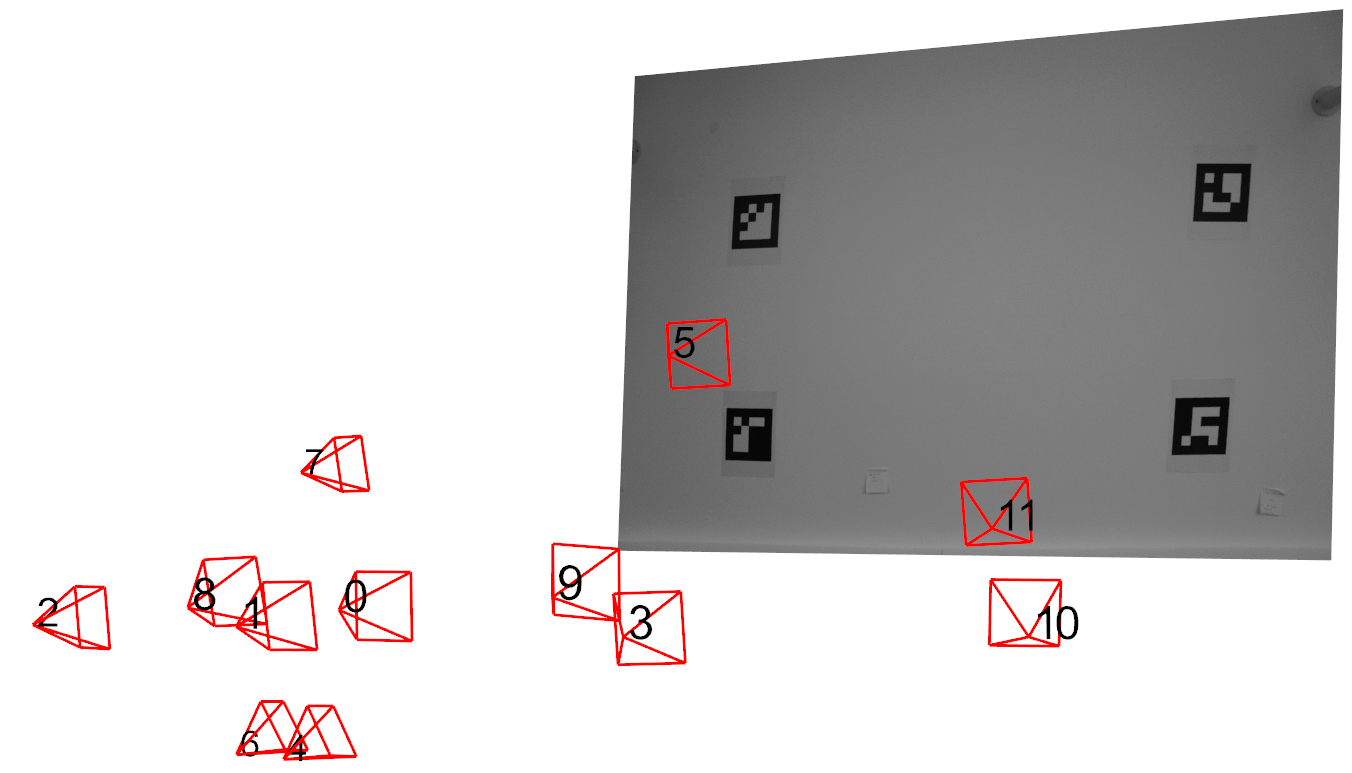}}
    \centerline{(b)}\medskip
	\end{minipage}
	\end{center}
		
	\caption{(a) Camera-light vision system for experiments. (b) Overview of test images.}
	\label{fig:experiment}
\end{figure}

\paragraph*{Setting} We built a camera-light vision system for evaluation by fixing a SONY Alpha 7 camera and a Bridgelux RS Array LED (with LEDiL CA12900 reflector) on a metal bar (see Figure \ref{fig:experiment} (a)). 
Twelve raw images were taken with this system from different views on a flat white wall (see Figure \ref{fig:experiment} (b)). 
One hundred valid pixels in each image are chosen for estimating the relative pose of the light.
The estimated relative pose of light and scale factor are solved by the procedure mentioned in the above Section \ref{sec:methodology}.
In our implementation, the optimization problem is solved by Ceres Solver \cite{ceres-solver}.

The evaluation is implemented on different numbers of images with the same very coarsely tape-measured initial values. The optimization is solved by using 12 to 8 images to evaluate the consistence of the method. 

\begin{figure}[htb]
	\begin{center}
		\begin{minipage}[a]{0.48\linewidth}
			\centering
			\centerline{\includegraphics[width=\linewidth]{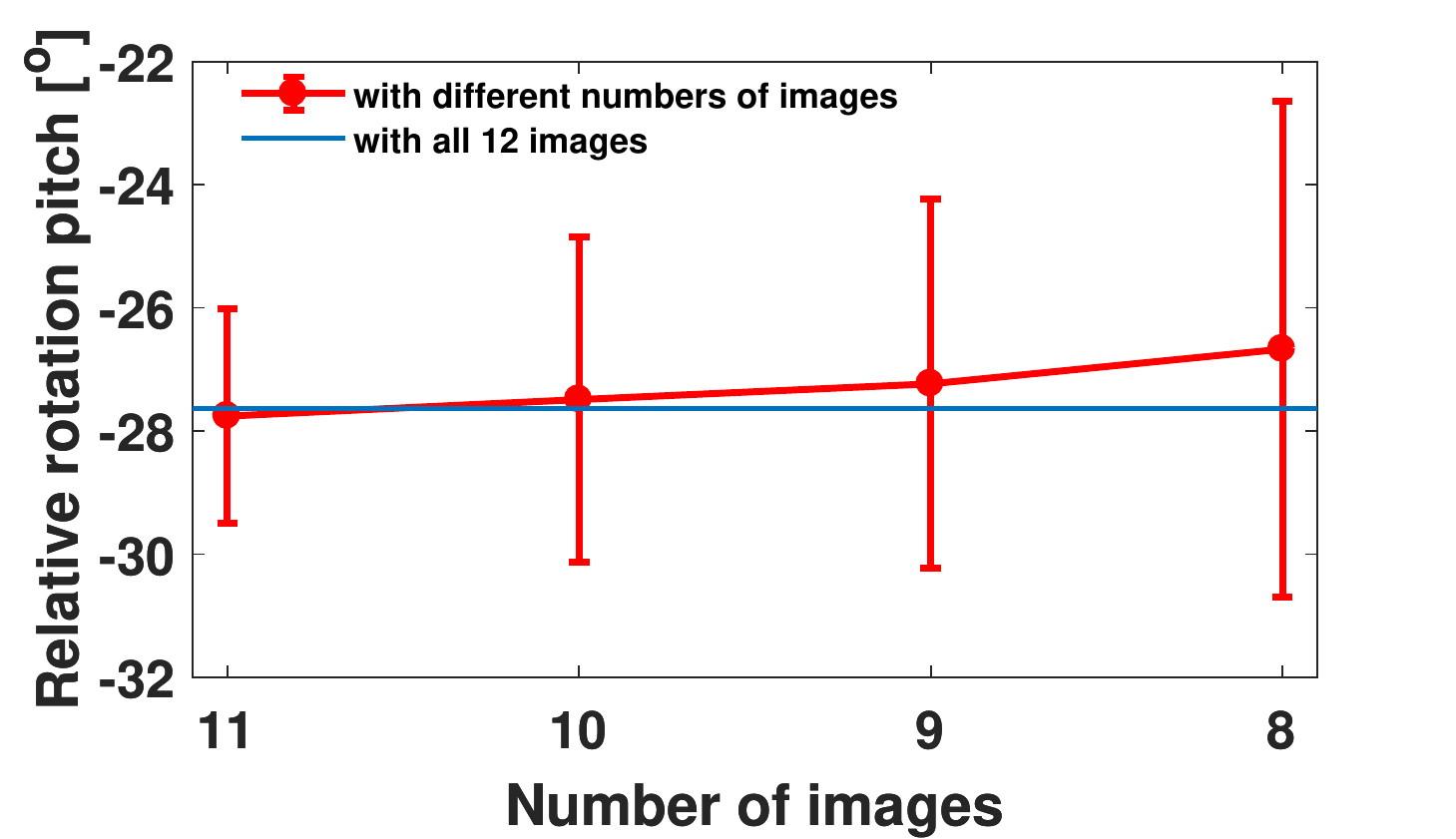}}
			\centerline{(a)}\medskip
		\end{minipage}
		\begin{minipage}[a]{0.48\linewidth}
			\centering
			\centerline{\includegraphics[width=\linewidth]{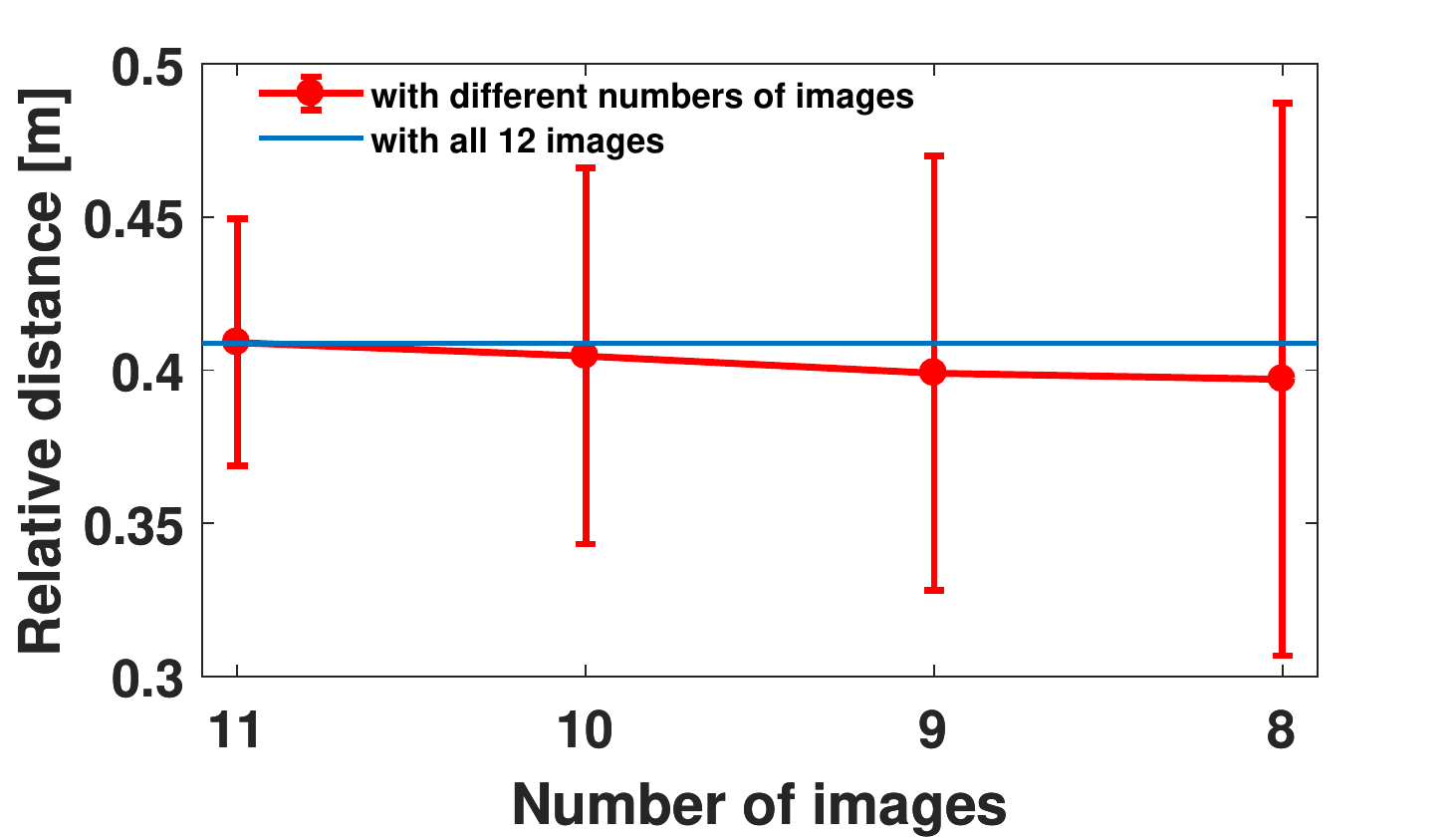}}
			\centerline{(b)}\medskip
		\end{minipage}
	\end{center}
	
	\caption{(a) Evaluation of estimated light orientation (Euler angle pitch) and (b) light position (converted to relative distance) with different numbers of input images.}
	\label{fig:evaluation}
\end{figure}

\paragraph*{Number of images}The results over different numbers of input images are shown in Figure \ref{fig:evaluation}, the estimation of relative pose of light is consistent and reliable. 
More input images yield more robust optimization results. 

\begin{figure}[t]
	\begin{center}
		\includegraphics[width=0.32\linewidth]{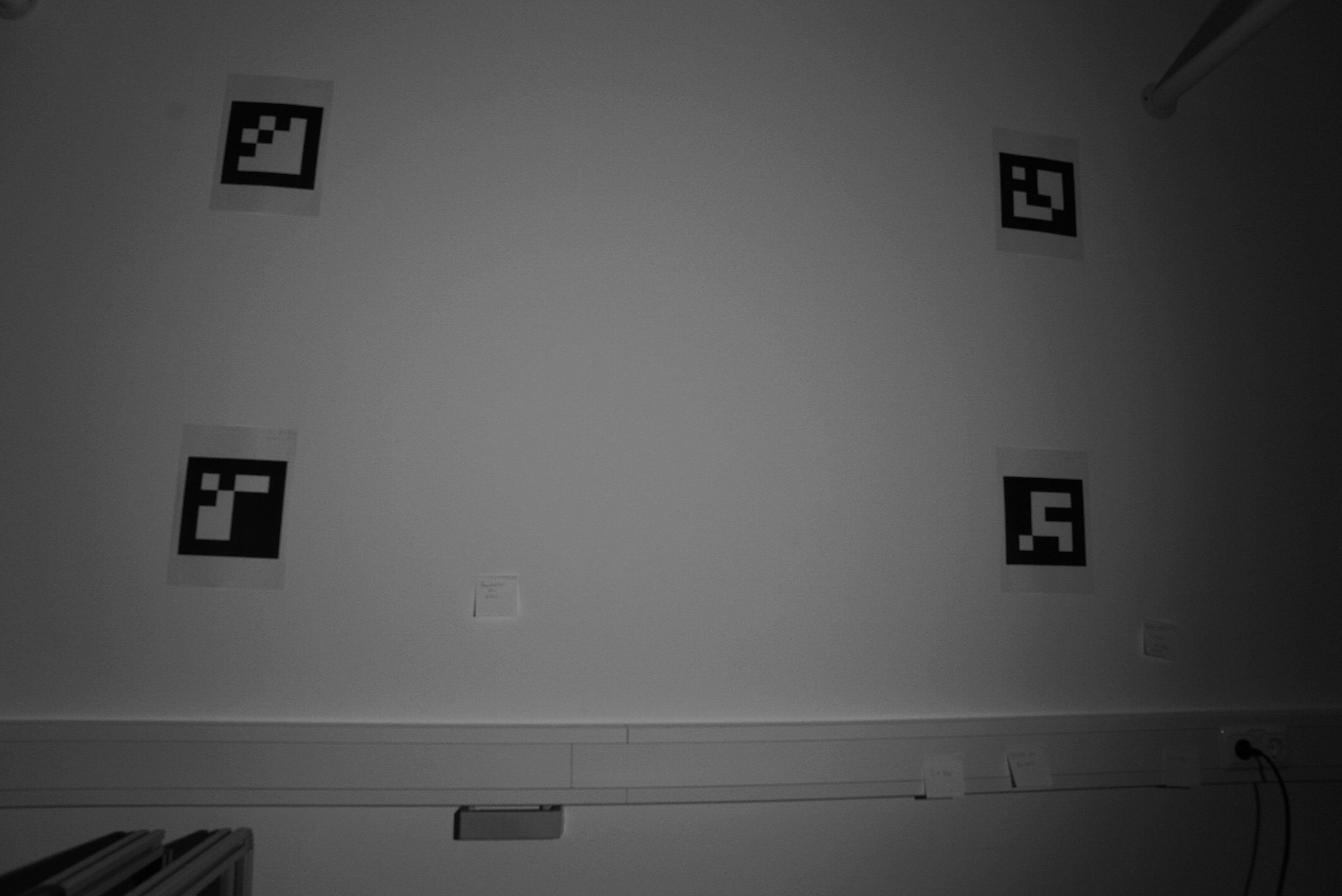}
		\hspace{0.05mm}
		\includegraphics[width=0.32\linewidth]{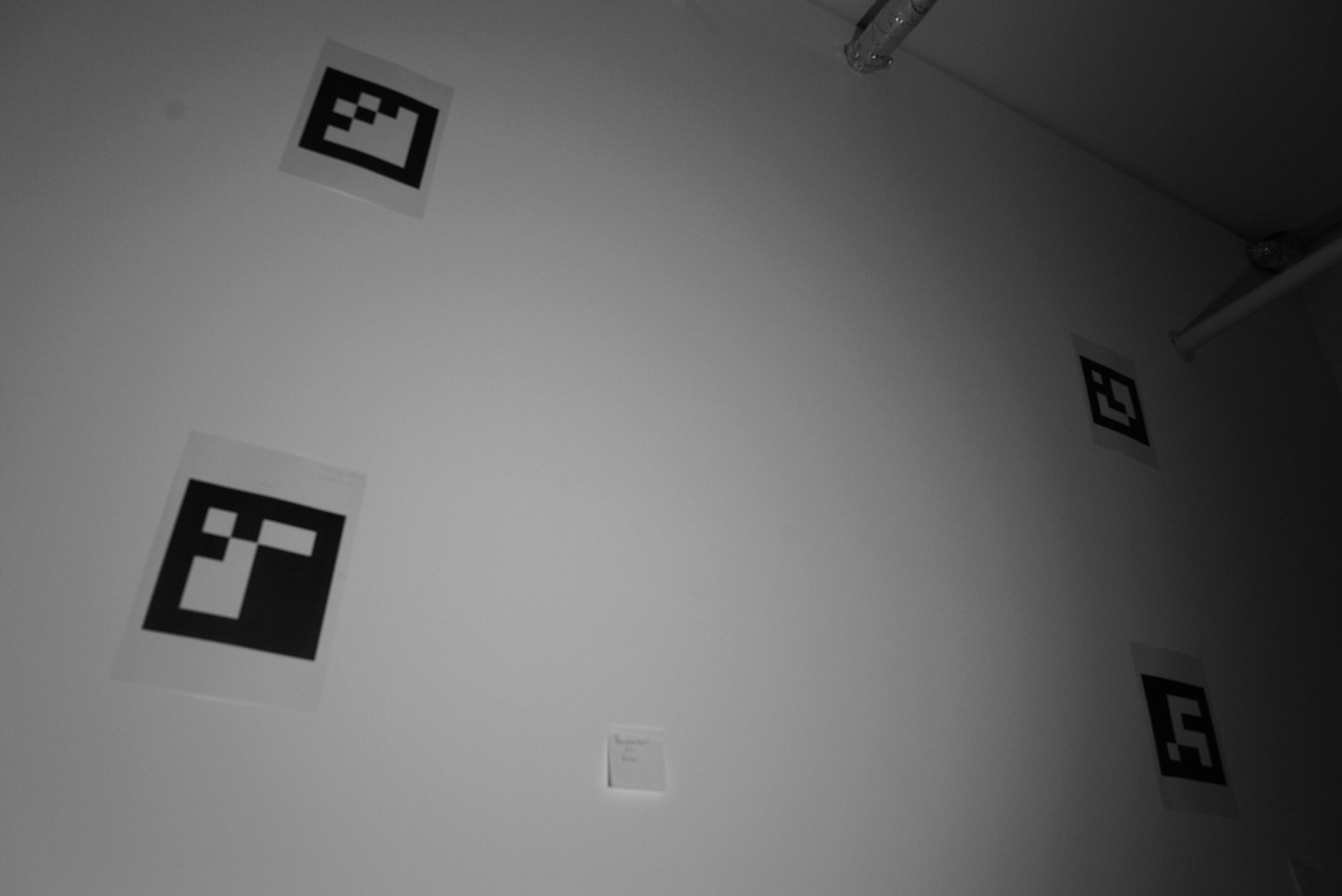}
		\hspace{0.05mm}
		\includegraphics[width=0.32\linewidth]{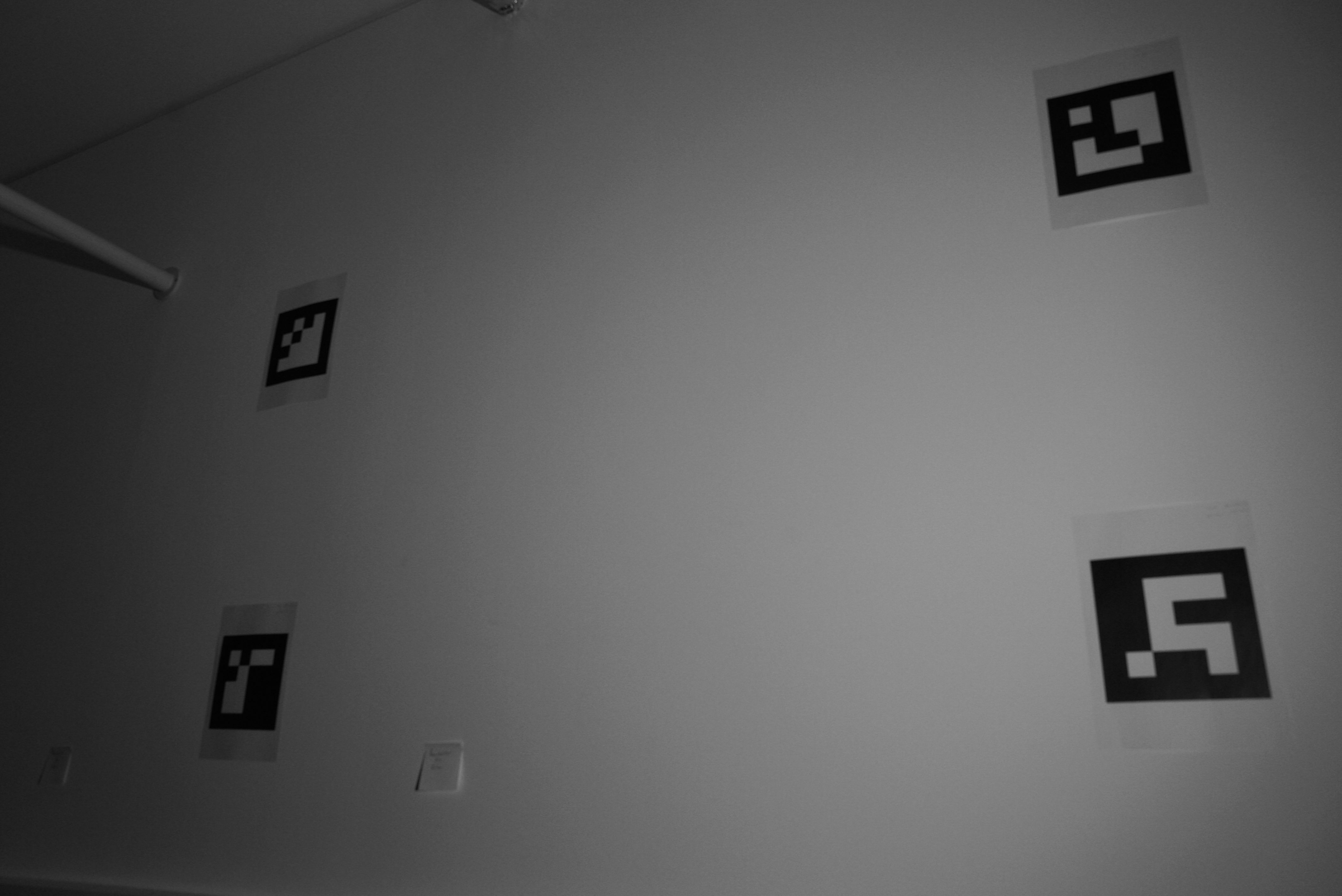}
		\hspace{0.05mm}
		\includegraphics[width=0.32\linewidth]{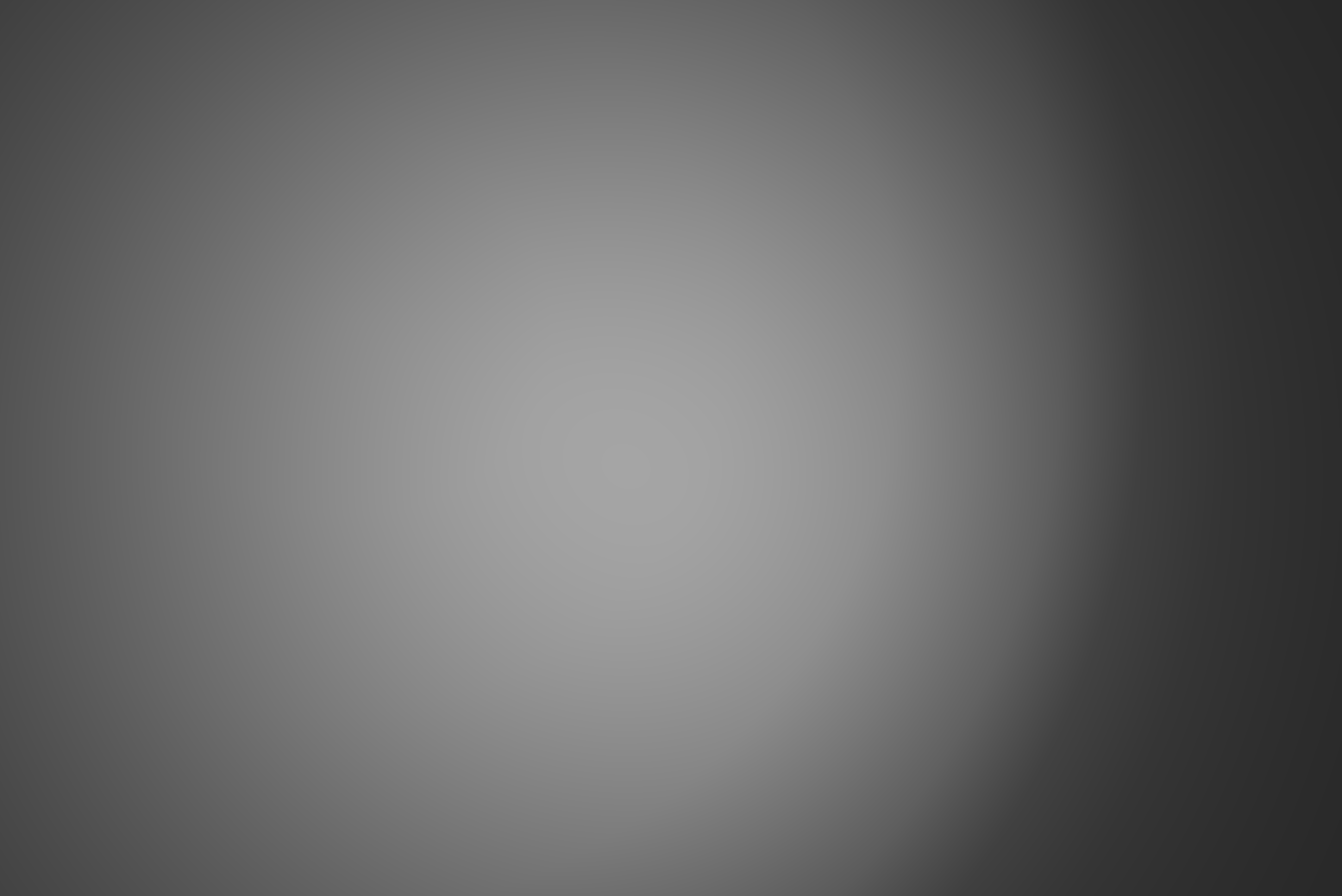}
		\hspace{0.05mm}
		\includegraphics[width=0.32\linewidth]{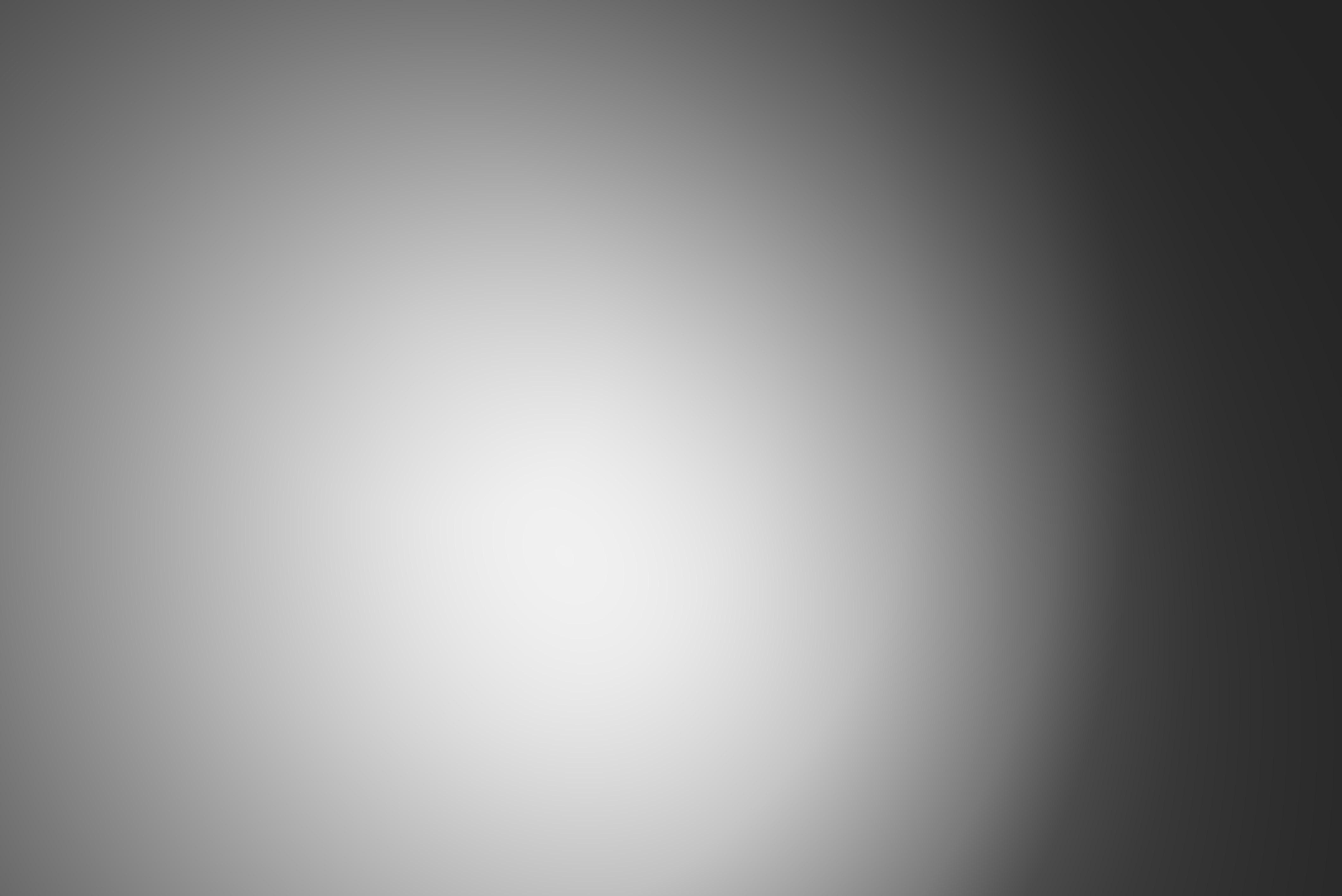}\
		\hspace{0.05mm}
		\includegraphics[width=0.32\linewidth]{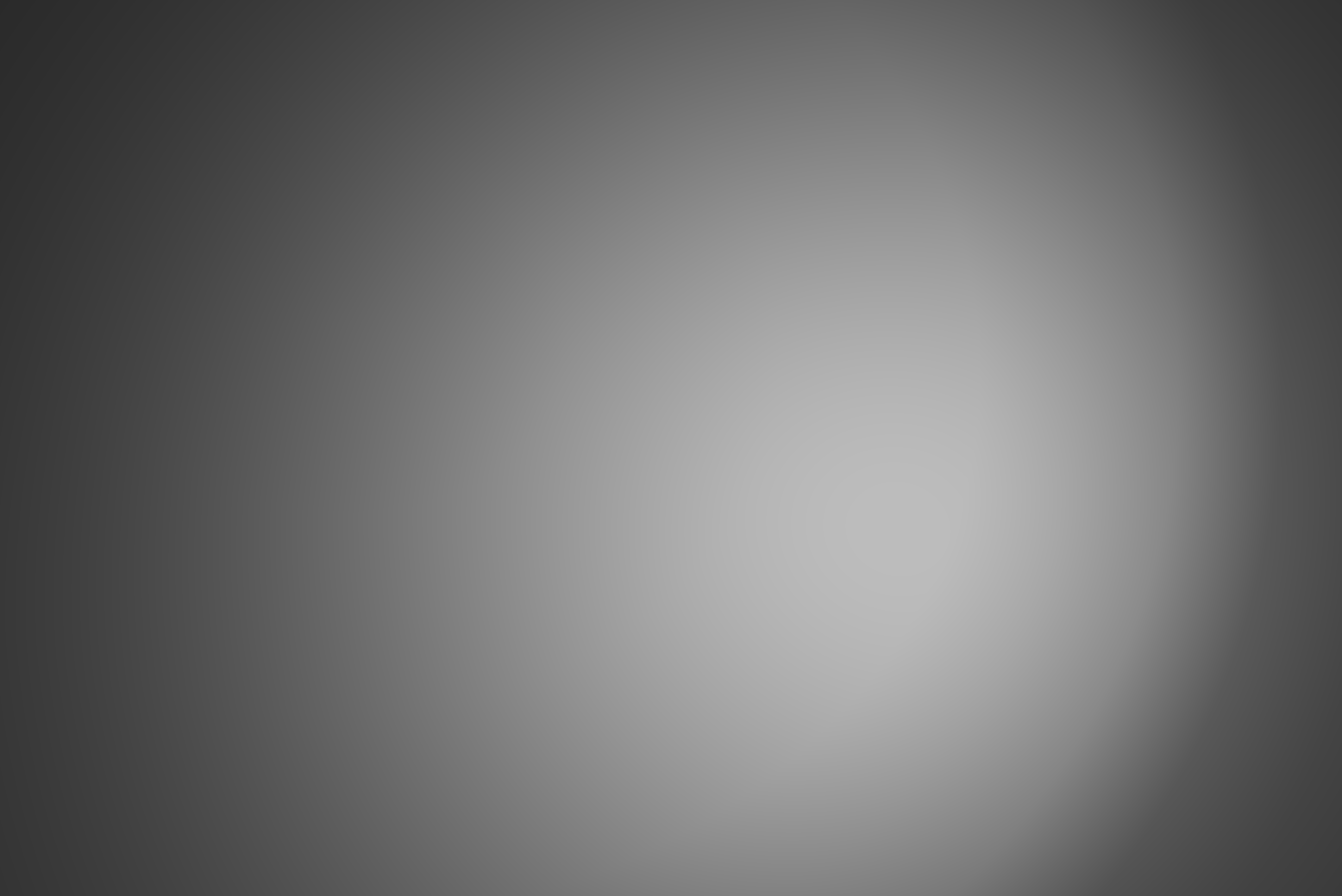}	
	\end{center}
	\caption{Upper row: real images. Bottom row: corresponding rendered images with the estimated light pose.}
	\label{fig:renderedImg}
\end{figure}

\paragraph*{Pose variation}
During the evaluation, we also noticed that the variation of distance and viewing angle will significantly improve the accuracy of light calibration results (similar to camera calibration). 
\paragraph*{Initialization}
An additional test with far off (errors $>$30$^{\circ}$ and $>$1m) initial values is also conducted. The calibration results still converged and remain consistent, which indicates that the optimization in the smooth, one-light setting has a large basin of convergence.
Full images are rendered in Figure \ref{fig:renderedImg} for giving more intuitive sense of the estimated light poses.

\section{Conclusions}
\label{sec:print}

This paper presents an optimization strategy, based on energy-preserving rendering for the calibration of the relative light pose wrt. the camera in a camera-light vision system. 
This method only requires to take several light pattern images on a flat reference plane from different views and distances as inputs. Upon that, the estimation of the light pose is solved by minimizing the residuals between real and rendered pixel intensities in an analysis-by-synthesis fashion, which is also suited to extend to multiple light source cases. 
The experimental results indicate that the method is able to estimate the relative light position and orientation consistently and robustly and independent of the initial value.
In the experiment, we applied the proposed method on a known symmetric non-isotropic light. However, this method is not limited by the types of lights and can be extended to other light types. Future work should examine how the light's radiance intensity distribution can best be included in the optimization 



%
%
%


\bibliographystyle{IEEEbib}

\end{document}